\documentclass[10pt,letterpaper]{article}

\usepackage{cogsci}

\cogscifinalcopy 

\usepackage{pslatex}
\usepackage{apacite}
\usepackage{float} 

\usepackage{tikz-dependency}
\usepackage{verbatim}
\usepackage{amsmath}
\usepackage{epsfig}
\usepackage{multirow}
\usepackage{linguex}
\usepackage{graphicx}
\usepackage{subfig}
\usepackage{devanagari}
\usepackage{wrapfig}
\usepackage{cases}
\usepackage{color}
\usepackage{booktabs}
\usepackage{nameref}
\usepackage[titletoc]{appendix}
\usepackage{mathtools}
\usepackage{lipsum}  
\usepackage{comment}
\usepackage{xurl}
\usepackage{commath}
\usepackage{gb4e}

\title{A bounded rationality account of dependency length minimization in Hindi}
 
\author{{\large \bf Sidharth Ranjan (sidharth.ranjan03@gmail.com)} \\
 University of Stuttgart, 70174 Stuttgart, Germany  \\
  \AND {\large \bf Titus von der Malsburg (titus.von-der-malsburg@ling.uni-stuttgart.de)} \\
  University of Stuttgart, 70174 Stuttgart, Germany}

\begin{document}

\maketitle

\begin{abstract}

  The principle of \textsc{dependency length minimization}, which seeks to keep syntactically related words close in a sentence, is thought to universally shape the structure of human languages for effective communication. However, the extent to which dependency length minimization is applied in human language systems is not yet fully understood. Preverbally, the placement of \textit{long-before-short} constituents and postverbally, \textit{short-before-long} constituents are known to minimize overall dependency length of a sentence. In this study, we test the hypothesis that placing only the shortest preverbal constituent next to the main-verb explains word order preferences in Hindi (a SOV language) as opposed to the global minimization of dependency length.  We characterize this approach as a least-effort strategy because it is a cost-effective way to shorten all dependencies between the verb and its preverbal dependencies. As such, this approach is consistent with the bounded-rationality perspective according to which decision making is governed by `fast but frugal’ heuristics rather than by a search for optimal solutions. Consistent with this idea, our results indicate that actual corpus sentences in the Hindi-Urdu Treebank corpus are better explained by the least effort strategy than by global minimization of dependency lengths. Additionally, for the task of distinguishing corpus sentences from counterfactual variants, we find that the dependency length and constituent length of the constituent closest to the main verb are much better predictors of whether a sentence appeared in the corpus than total dependency length. Overall, our findings suggest that cognitive resource constraints play a crucial role in shaping natural languages.

\end{abstract}

\section{Introduction}\label{sect:intro}

Human working memory is viewed as a limited capacity system, where there is pressure to minimize the cognitive load by not retaining information longer than necessary. \citeA{simon1982models, simon1990invariants, simon1991cognitive} proposed the idea of \textit{bounded rationality} in decision making which incorporates the aforementioned cognitive limitation and proposed that human's ability to make rational decisions has adapted to making choices that are satisfactory rather than optimal.  In language, speakers’ decisions are influenced by the context in which the decision is made---\textit{availability of information, cognitive resources,} and the \textit{limited response time}.  This tension naturally exerts a challenge for speakers and constrains them to adopt certain strategies to formulate sentences that are most efficient for communication.  For example, choosing word-order patterns that are most common in daily discourse or placing the words/phrases early in the sentence that are more accessible or salient in working memory.  In this work, we examine whether word order variation in Hindi can be seen as a result of the pressures for efficient communication within the constraints of bounded rationality.

Prior work has investigated the role of phrase length on the constituent ordering of sentences~\cite{Bock1985, Arnold00heaviness, hawkins1994, yamashitaChang2001, Choi2007, faghiri2020word}. The findings suggest that SVO languages (e.g., English) have a preference for \textit{short-before-long} linguistic structure such that shorter phrases are more accessible and get realized before relatively less accessible or longer ones~\cite{Arnold00heaviness}. In contrast, for SOV languages (e.g., Japanese), \textit{long-before-short} ordering is prevalent, \textit{i.e.,} longer phrases tend to be shifted ahead of shorter ones with a view that long phrases are semantically rich and conceptually salient, and thus realized before than the shorter ones~\cite{yamashitaChang2001}. In this vein, Gibson’s  Dependency Locality Theory provides a unified explanation for constituent ordering patterns in both SOV and SVO languages \cite{gibson00}. The theory prefers constituent order that minimizes the overall dependency length of the sentence irrespective of language. Preverbally, \textit{long-before-short} and postverbally, \textit{short-before-long} constituent ordering minimize the overall dependency length in the sentence~\cite{hawkins04}.

The principle of \textsc{dependency length minimization} (DLM), is believed to reduce the cognitive load on working memory by reducing the distance between syntactically related words in the sentence for both speakers and hearers~\cite{gibson1991computational,hudson1995measuring,Gib98,gibson00,i2004euclidean,Temperley2007}. This has been shown to be the case in natural languages, which tend to have shorter overall dependency lengths than those produced randomly~\cite{futrell2015,Liu2017,temperley-gildea-ar18,futrell2020dependency}. The DLM hypothesis has therefore been demonstrated to be an essential characteristic of efficient word order by virtue of communicative efficiency in human language~\cite{Gibson2019}. However, the extent to which the minimization of dependency length is applied to human language systems in not yet fully understood. This serves as our primary motivation, for which we apply the idea of bounded rationality to DLM and investigate how well it explains word-order preferences in Hindi.

\newpage

\begin{scriptsize}
\begin{exe}

  \ex \label{ex:hindi-intro}
  \begin{xlist}
  \ex[]{\label{ex:h1}
    \gll {\bf \color{red}{maa=\textbf{ne}}} {\bf \color{cyan}{baajaar jaate samaye}} {\bf \color{violet}{rote hue bacche=\textbf{ko}}} {\bf toffee} {di}\\
         {mother=\textsc{erg}} {market ~~going while} {cry=\textsc{prog} child=\textsc{acc}} {toffee} {di-\textsc{pfv.f.sg}}\\
     \glt \textit{The mother gave the crying child a candy while going to market.}}

  \ex[] {\label{ex:h2}  {\bf \color{red}{maa=\textbf{ne}}} {\bf \color{violet}{rote hue bacche=\textbf{ko}}} {\bf \color{cyan}{baajaar jaate samaye}} {\bf toffee} {di}}
  \ex[] {\label{ex:h3} {\bf \color{violet}{rote hue bacche=\textbf{ko}}} {\bf \color{cyan}{baajaar jaate samaye}} {\bf \color{red}{maa=\textbf{ne}}} {\bf toffee} {di}}

  \end{xlist}

\end{exe}
\end{scriptsize}

Hindi, a Indo-Aryan language from the Indo-European language family, has subject-object-verb (\textsc{sov}) as the basic word order and has a rich case-marking system~\cite{kachru1982}. In contrast to English, Hindi exhibits greater flexibility in the arrangement of words within a sentence, as illustrated in Example \ref{ex:hindi-intro}. Example \ref{ex:h1} has 4 preverbal constituents and is considered the most preferred syntactic choice since it originally appeared in the corpus as opposed to all variants (4! = 24) that are possible with this sentence.

In this work, we test our hypothesis that speakers adopt a least-effort strategy to determine the word order choices in Hindi as opposed to global minimization of dependency length.  We define \textit{least-effort strategy} as the placement of shortest preverbal constituent closest to the main verb as this is the most economical way of reducing all the dependencies between the verb and its preverbal dependents (see Figure \ref{fig:ordering-strategies} for an illustration). We generated different counterfactual variants by randomly rearranging their preverbal constituents (see Examples \ref{ex:h2}-\ref{ex:h3} to list a few) for each reference sentence (\ref{ex:h1}) in Hindi-Urdu Treebank corpus of written text. Next, we conducted a quantitative analysis of those variants and their corresponding corpus reference sentences. We then deployed a logistic regression classifier to distinguish the corpus reference sentence from the artificially generated variants based on the dependency length and constituent length of preverbal constituents, and overall dependency length of the sentence. The primary motivation behind choosing constituent-level predictors was to test the efficacy of the least-effort strategy (see Figure \ref{ex:least-effort-order}) against global minimization constraint (see Figure \ref{ex:descending-order}) under the purview of bounded rationality. As the number of preverbal constituents increases (say from 3 $\rightarrow$ 5), the number of possible variants also increases significantly (3! = 6 $\rightarrow$ 5! = 120); overburdening the speakers to make choices. A boundedly rational speaker would then apply a heuristic strategy that identifies a satisfactory solution that does not necessarily minimize total dependency length \cite{NewellSimon1972a,gigerenzer2011heuristic}. 

Our main contribution through this work is that we show the impact of dependency length minimization on word order choices in Hindi using large-scale naturalistic data rather than the constructed experimental stimuli and offer cross-linguistic evidence essential for the advancement of cognitive theories~\cite{Jaeger2009compass}. 

The remainder of this paper is structured as follows. We first briefly describe the data and methods used to investigate the hypothesis, followed by the results in this work. Towards the end, we discuss the broader implications of our findings with a conclusion and future directions.

\section{Method}\label{sect:method}

\begin{figure*}%
    \centering
    \subfloat[\centering Reference sentences (7586 data points) \label{ref-const-counts}]{{\includegraphics[trim={0cm 0 1cm 2cm}, clip, scale = 0.37]{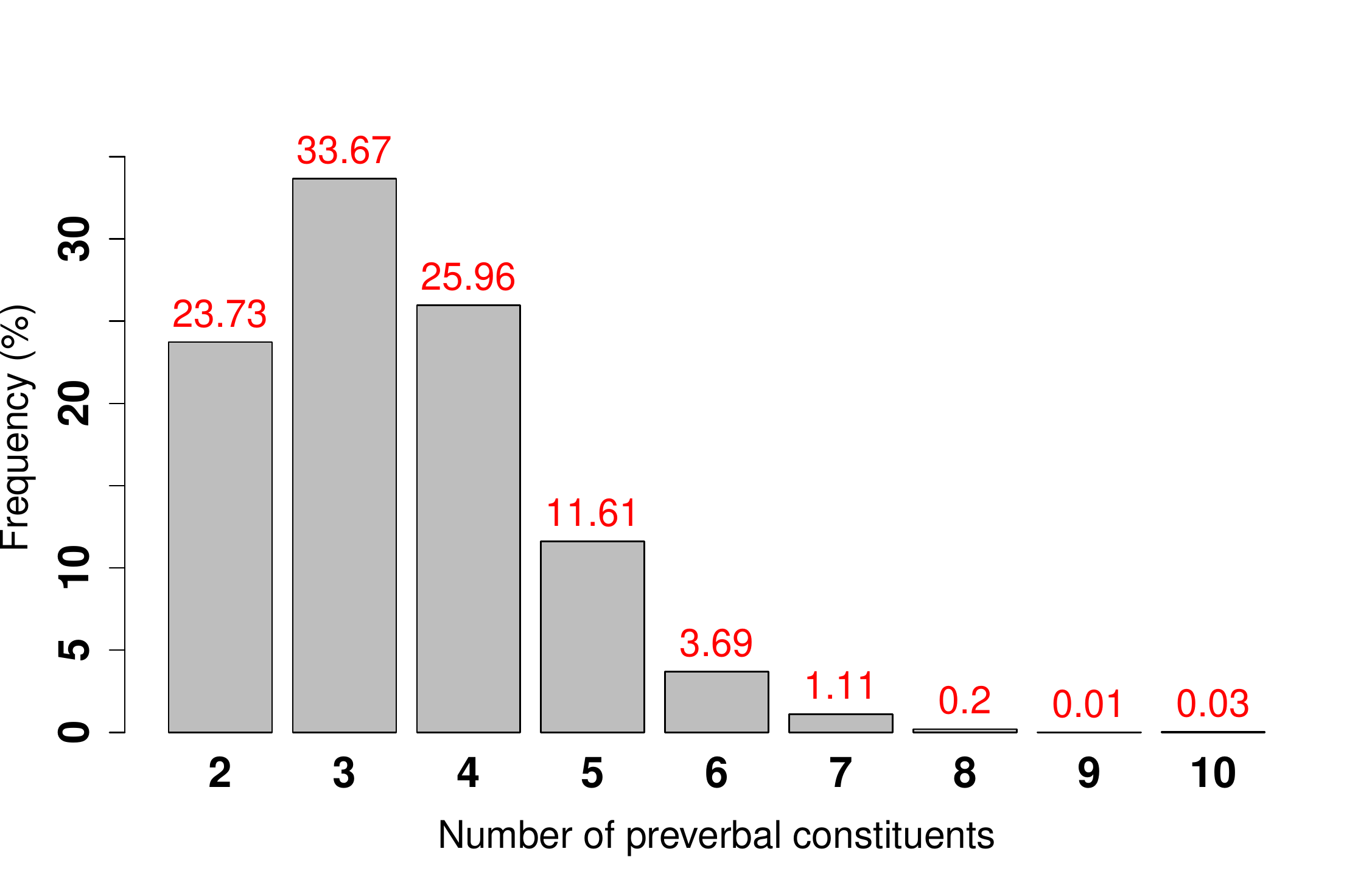} }}%
    \qquad
    \subfloat[\centering Variant sentences (184818 data points) \label{var-const-counts}]{{\includegraphics[trim={0cm 0 1cm 2cm}, clip, scale=0.37]{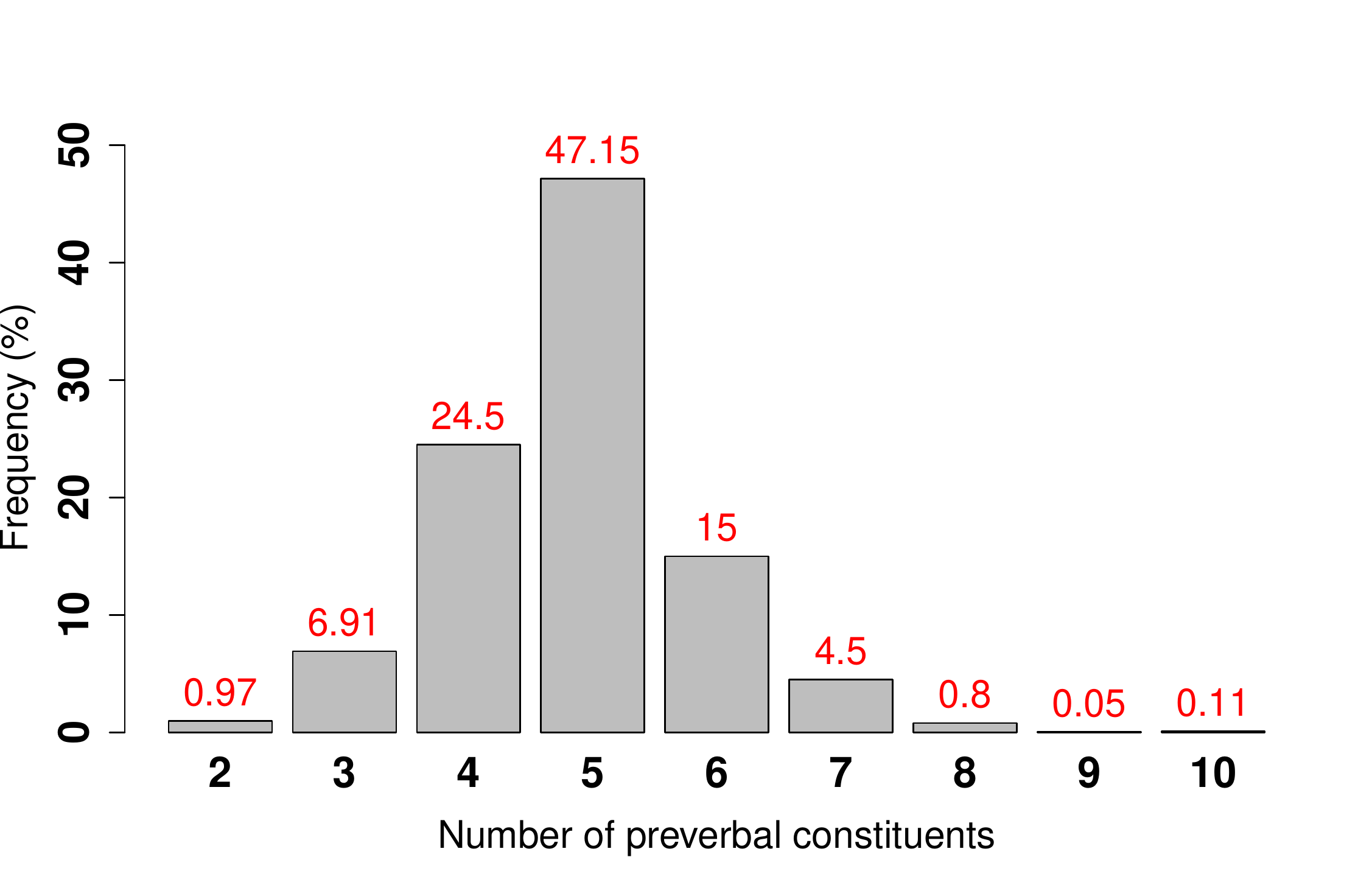} }}%
    \caption{Percentages of reference and variant sentences with varying preverbal constituents}%
    \label{fig:preverbal-const-count}%
\end{figure*}

Our data set consists of 7,586 declarative sentences from the Hindi-Urdu Treebank (HUTB) corpus of written text belonging to the newswire domain~\cite{Bhatt2009}. For each reference sentence in the corpus, we created artificial variants by randomly permuting the preverbal constituents whose head was immediate child to the root verb in their projective dependency tree (see Figure \ref{fig:ordering-strategies}). In instances where there were more than 100 variants (an arbitrary cutoff\footnote{Higher or lower cutoffs do not impact our results.} to keep our computation tractable), we randomly selected 99 of them. Our variant generation approach resulted in 184818 competing variants in total, for the reference sentences in our dataset. Refer to Example \ref{ex:hindi-intro} for an illustration. Figure \ref{fig:preverbal-const-count} depicts the percentages of reference-variant sentences for each number of preverbal constituents. Due to insufficient data points for sentences with 7 or more preverbal constituents, we only present the corpus analysis results for sentences with up to 6 preverbal phrases. However, for the binary classification task presented in the later section of this work, we used the entire dataset to compute our results \footnote{We also carried out a similar analysis with only grammatical variants and found that our findings were consistent with those reported below.}.

\begin{figure}
    \centering
    \includegraphics[scale=0.24,trim={10mm 1mm 35mm 1mm},clip]{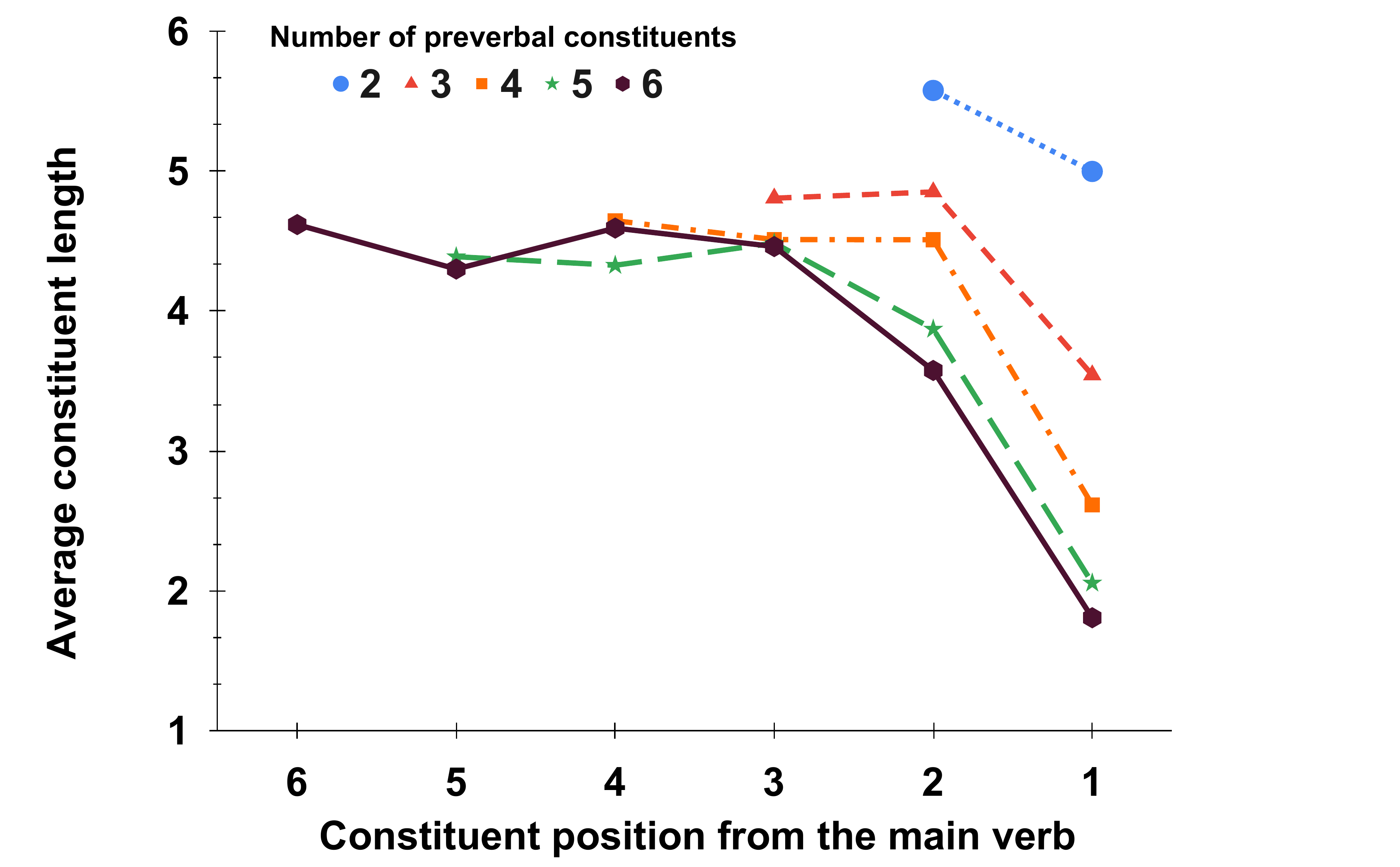}
    \caption{Average constituent length of preverbal constituents for corpus sentences with 2 to 6 constituents separately}
    \label{fig:preverbal-constlen}
\end{figure}

\section{Corpus analysis}\label{subsect:corpus-analysis}

Previous corpus studies in Hindi have revealed a predominance of \textit{long-before-short} constituent orders preverbally, and smaller head-dependent distances~\cite{sharma2020determines,cog:sid}. To test our hypothesis of least-effort strategy, we investigated the role of all preverbal constituents in minimizing the overall dependency length of the sentence. Consistent with previous studies, we also found that \textit{long-before-short} constituent orders are dominant in Hindi~\cite{cog:sid}. In addition, we also observed a wide distribution of \textit{long-before-short} and \textit{short-before-long} constituents in corpus sentences with varying preverbal constituents (see Figure~\ref{fig:preverbal-constlen}). Crucially, the plot in Figure \ref{fig:preverbal-constlen} suggests the constituent closest to the main verb has a strong tendency to be the shortest among all the preverbal constituents in the sentence. This is consistent with the predictions of the least effort principle, because moving the shortest constituent to the main verb minimizes not just that constituent's dependency length but also the dependency lengths of all other preverbal dependents of the main verb, therefore striking a favorable cost-benefit ratio. 

\begin{figure*}[t]
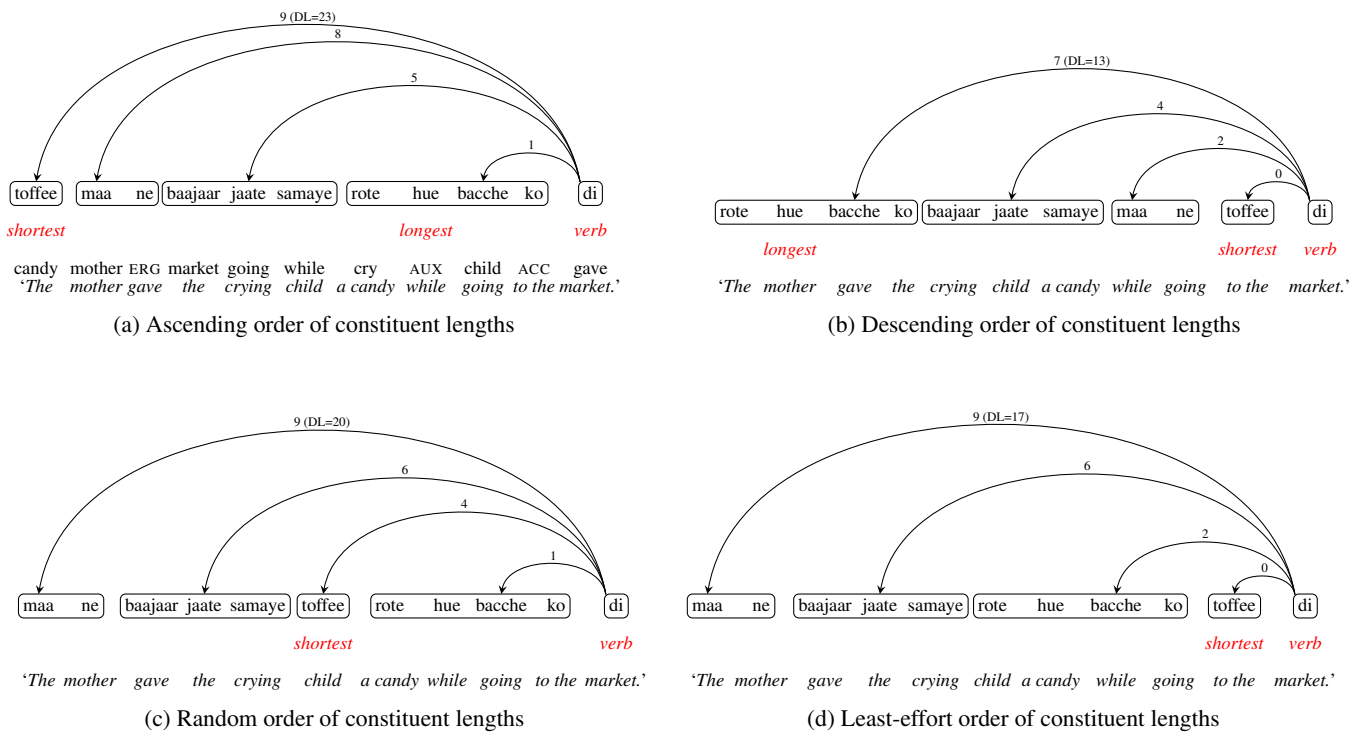

\noindent\makebox[\textwidth]{%
\subfloat[Subfigure 3 list of tables text][Ascending order of constituent lengths]{
\label{ex:ascending-order}
\begin{scriptsize}
\begin{dependency}[arc edge, arc angle=80, text only label, label style={above}]
\begin{deptext}[column sep=-0.05cm]
toffee \& maa \& ne \& baajaar \& jaate \& samaye \& rote \& hue \& bacche \& ko \& di\\
\& \& \& \& \& \& \& \& \& \& \\
\textcolor{red}{\it shortest}\& \& \& \& \& \& \& \textcolor{red}{\it longest} \& \& \& \textcolor{red}{\it verb}\\
\& \& \& \& \& \& \& \& \& \& \\
candy \& mother \& \textsc{erg}  \& market \& going \& while \& cry \& \textsc{aux} \& child \& \textsc{acc} \& gave\\
`\textit{The} \& \textit{mother} \& \textit{gave} \& \textit{the} \&  \textit{crying} \& \textit{child} \& \textit{a candy} \& \textit{while} \& \textit{going} \& \textit{to the} \& \textit{market.}'\\
\end{deptext}
\depedge{11}{1}{9 (DL=23)}
\depedge{11}{2}{8}
\depedge{11}{5}{5}
\depedge{11}{9}{1}
\wordgroup{1}{1}{1}{subj}
\wordgroup{1}{2}{3}{obj1}
\wordgroup{1}{4}{6}{pp1}
\wordgroup{1}{7}{10}{verb}
\wordgroup{1}{11}{11}{verb}
\end{dependency}
\end{scriptsize}
}
\qquad
\subfloat[Subfigure 4 list of tables text][Descending order of constituent lengths]{
\label{ex:descending-order}
\begin{scriptsize}

\begin{dependency}[arc edge, arc angle=80, text only label, label style={above}]
\begin{deptext}[column sep=0.05cm]
rote \& hue \& bacche \& ko \& baajaar \& jaate \& samaye \& maa \& ne \& toffee \& di\\
\& \& \& \& \& \& \& \& \& \& \\
\& \textcolor{red}{\it longest} \& \& \& \& \& \& \& \& \textcolor{red}{\it shortest} \& \textcolor{red}{\it verb}\\
\& \& \& \& \& \& \& \& \& \& \\
`\textit{The} \& \textit{mother} \& \textit{gave} \& \textit{the} \&  \textit{crying} \& \textit{child} \& \textit{a candy} \& \textit{while} \& \textit{going} \& \textit{to the} \& \textit{market.}'\\
\end{deptext}
\depedge{11}{3}{7 (DL=13)}
\depedge{11}{6}{4}
\depedge{11}{8}{2}
\depedge{11}{10}{0}
\wordgroup{1}{1}{4}{subj}
\wordgroup{1}{5}{7}{obj1}
\wordgroup{1}{8}{9}{pp1}
\wordgroup{1}{10}{10}{verb}
\wordgroup{1}{11}{11}{verb}
\end{dependency}
\end{scriptsize}
}
}
\\
\noindent\makebox[\textwidth]{%
\subfloat[Subfigure 1 list of tables text][Random order of constituent lengths]{
\label{ex:random-order}
\begin{scriptsize}
\begin{dependency}[arc edge, arc angle=80, text only label, label style={above}]
\begin{deptext}[column sep=0cm]
maa \& ne \& baajaar \& jaate \& samaye \& toffee \& rote \& hue \& bacche \& ko \& di\\
\& \& \& \& \& \& \& \& \& \& \\
\& \& \& \& \& \textcolor{red}{\it shortest} \& \& \& \& \& \textcolor{red}{\it verb}\\
\& \& \& \& \& \& \& \& \& \& \\
`\textit{The} \& \textit{mother} \& \textit{gave} \& \textit{the} \&  \textit{crying} \& \textit{child} \& \textit{a candy} \& \textit{while} \& \textit{going} \& \textit{to the} \& \textit{market.}'\\
\end{deptext}
\depedge{11}{1}{9 (DL=20)}
\depedge{11}{4}{6}
\depedge{11}{6}{4}
\depedge{11}{9}{1}
\wordgroup{1}{1}{2}{subj}
\wordgroup{1}{3}{5}{obj1}
\wordgroup{1}{6}{6}{pp1}
\wordgroup{1}{7}{10}{verb}
\wordgroup{1}{11}{11}{verb}
\end{dependency}
\end{scriptsize}
}
\subfloat[Subfigure 2 list of tables text][Least-effort order of constituent lengths]{
\label{ex:least-effort-order}
\begin{scriptsize}
\begin{dependency}[arc edge, arc angle=80, text only label, label style={above}]
\begin{deptext}[column sep=0.03cm]
maa \& ne \& baajaar \& jaate \& samaye \& rote \& hue \& bacche \& ko \& toffee \& di\\
\& \& \& \& \& \& \& \& \& \& \\
\& \& \& \& \& \& \& \& \& \textcolor{red}{\it shortest} \& \textcolor{red}{\it verb}\\
\& \& \& \& \& \& \& \& \& \& \\
`\textit{The} \& \textit{mother} \& \textit{gave} \& \textit{the} \&  \textit{crying} \& \textit{child} \& \textit{a candy} \& \textit{while} \& \textit{going} \& \textit{to the} \& \textit{market.}'\\
\end{deptext}
\depedge{11}{1}{9 (DL=17)}
\depedge{11}{4}{6}
\depedge{11}{8}{2}
\depedge{11}{10}{0}
\wordgroup{1}{1}{2}{subj}
\wordgroup{1}{3}{5}{obj1}
\wordgroup{1}{6}{9}{pp1}
\wordgroup{1}{10}{10}{verb}
\wordgroup{1}{11}{11}{verb}
\end{dependency}
\end{scriptsize}
}
}
\caption{Dependency length and constituent ordering for Hindi head-final structure; Only main verb dependencies are depicted in the figures; Total dependency length (DL) of the structure indicated above each sub-figure; Constituent dependency length is mentioned above each dependency arc}
\label{fig:ordering-strategies}
\end{figure*}

Interestingly, Figure~\ref{fig:preverbal-constlen} also shows that the pressure to move shorter constituent to the main verb increases with the number of preverbal constituents, thus suggesting that speakers may dynamically balance production cost and comprehenders' cognitive constraints.\footnote{Pearson's correlation coefficient between sentence length and number of preverbal constituents was 0.45.} To further test support the idea that speakers adopt a least-effort strategy (i.e. place shortest constituent next to the main verb), we compared the total dependency length of reference sentences (attested in the corpus) to four alternatives with different constituent orders illustrated in Figure \ref{fig:ordering-strategies} for example sentence \ref{ex:h1}:

\begin{small}
\begin{enumerate}
    \item \textsc{Ascending order:} Arrange the preverbal constituents in their increasing order of constituent lengths. This arrangement leads to maximal dependency length of the sentence (See Figure \ref{ex:ascending-order}).
    \item \textsc{Descending order:} Arrange the preverbal constituents in their decreasing order of constituent lengths. This arrangement globally minimizes the dependency length of the sentence (See Figure \ref{ex:descending-order}).
    \item \textsc{Random order:} Arrange the preverbal constituents randomly (See Figure \ref{ex:random-order}).
    \item \textsc{Least-effort order:} Start with a random order of preverbal constituents and then simply move the shortest constituent next to the main verb (See Figure \ref{ex:least-effort-order}).
\end{enumerate}
\end{small}

\begin{figure*}
    \centering
    \includegraphics[scale=0.48,trim={2mm 2mm 0mm 1mm},clip]{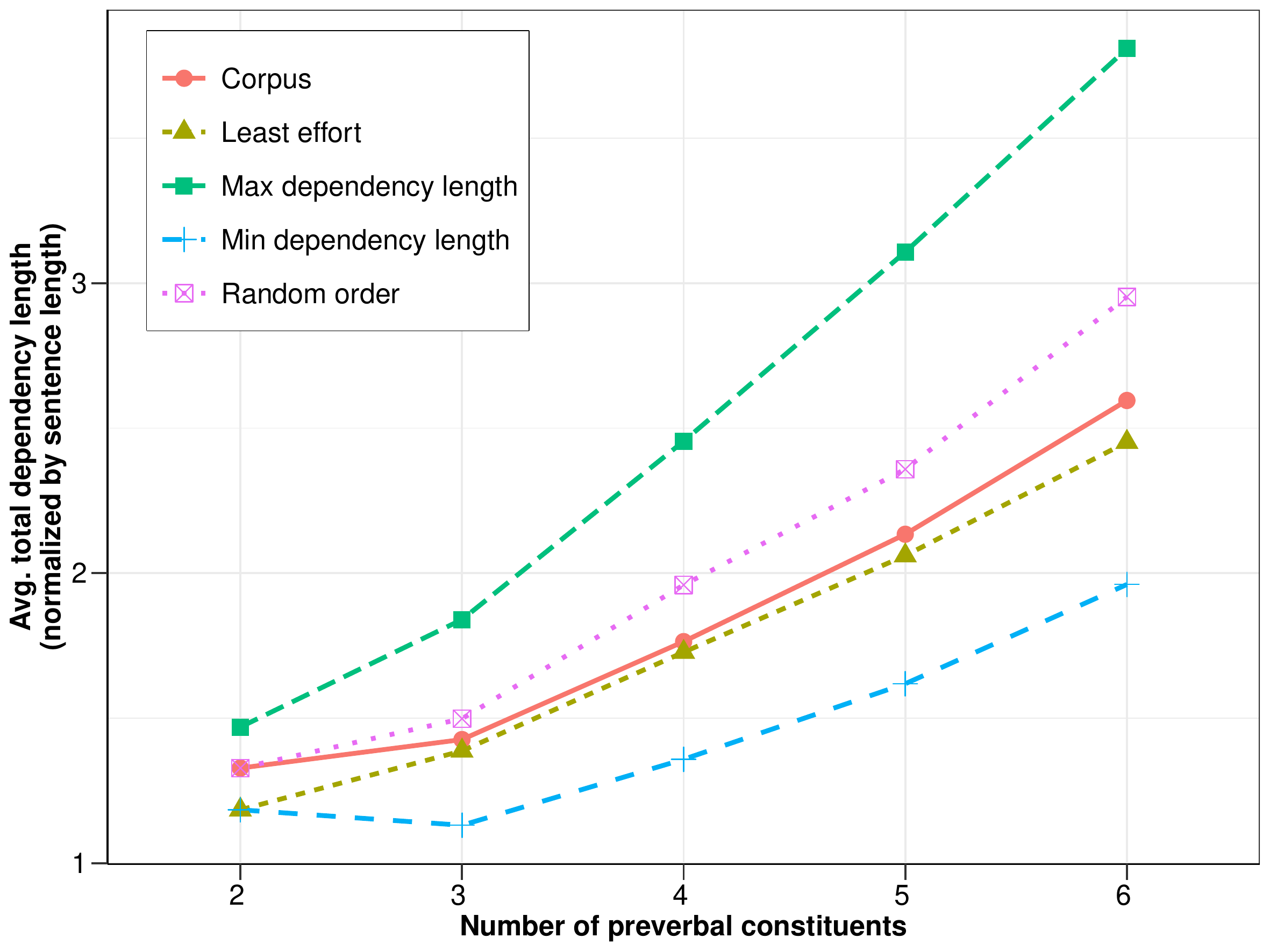}
    \caption{Average total dependency length (\textit{i.e.,} total dependency length normalized by the number of words in a sentence) for different constituent orderings and for different numbers of preverbal constituents (7586 data points for each ordering)}
    \label{fig:bounded-rationality}
\end{figure*}

Our results in Figure \ref{fig:bounded-rationality} show that the dependency length of sentences in the corpus generally tracks the dependency length of the least-effort solution for sentences with three or more preverbal constituents.  Only sentences with two preverbal constituents deviate from the least-effort order.  This may be because the memory pressure is so low in these cases that the system does not optimize dependency length at all, not even with the least effort heuristic.  Consistent with previous studies, the plot also confirms that corpus reference sentences on average have lower dependency length than the sentences with random word orders~\cite{Liu2008,GildeaT10,futrell2015,futrell2020dependency}.

These results further corroborate the idea that dependency length minimization in Hindi, and thus constituent order, are governed by a least-effort strategy.  This is consistent with the bounded rationality view of decision making, based on which we would expect that speakers consider only a limited search space when planning sentences~\cite{gigerenzer2011heuristicsbook}.

In the next section, we further test and validate the least-effort hypothesis using a decision task with a goal to distinguish reference sentences from random variants~\cite{GildeaT10,temperley-gildea-ar18}.

\section{Computational simulation}\label{sect:results}

We set up a binary classification task to distinguish the reference sentence amidst the artificially generated variants. Our original dataset had a significant class imbalance, with 184818 variants compared to only 7586 reference sentences. To address this imbalance, we applied the technique suggested by \citeA{Joachims:2002} which transforms the task of classifying a sentence as either reference or variant into a task of ranking reference sentence against each of its variants (pairwise ranking) by training the classifier on the difference between the feature vectors of variant sentence and corresponding paired reference sentence (see Equations \ref{eq:nor} and \ref{eq:joc}).

\begin{small}
\begin{equation}\label{eq:nor}
 w~\cdot~\phi(reference) > w~\cdot~\phi(variant)
\end{equation}
\end{small}

\vspace{-1em}

\begin{small}
\begin{equation}\label{eq:joc}
 w~\cdot~(\phi(reference)~-~\phi(variant)) > 0    
\end{equation}
\end{small}

Equation \ref{eq:nor} represents the objective of a standard binary classifier where corpus reference sentence is preferred against its corresponding variant sentence. The classifier must determine a feature weight ($w$) such that the dot product of $w$ and the reference feature vector ($\phi(reference)$) is greater than the dot product of $w$ and the variant feature vector ($\phi(variant)$). This objective can be rephrased as Equation \ref{eq:joc} where the dot product of $w$ and the difference between the feature vectors must be greater than zero. The dataset was structured such that each variant sentence was paired with its corresponding reference sentence, with the order being balanced across these pairs. For instance, Example \ref{ex:hindi-intro} would give (\ref{ex:h1}-\ref{ex:h2}) and (\ref{ex:h3}-\ref{ex:h1}). Feature vectors were subtracted and binary labels were assigned to each transformed data point. The ``Reference-Variant" pairs were labeled as ``1" and ``Variant-Reference" pairs were labeled as ``0", thus balancing the previously imbalanced classification task. Post \citeauthor{Joachims:2002}'s transformation, we had in total 184818 data points for our classification task. See \citeA{cog:sid} for more details.

We used the \texttt{glm} function in R to test the stated hypothesis at the outset (dependent variable $\sim$ independent variables): $choice \sim  \delta~dependency~length$ and $choice \sim  \delta~constituent~length$. Here, choice is a binary choice dependent variable (1 stands for reference sentence preference, and 0 denotes the variant sentence preference). The delta ($\delta$) refers to the difference between the feature vectors of reference sentence and its paired variant. The classifier predicts label ``1" if corpus reference sentence outranks a variant paired with it and ``0" if variant outranks its paired reference sentence. All the independent variables were normalised to $z$-scores, {\em i.e.}, the predictor's value (centered around its mean) was divided by its standard deviation. We estimated independent variables as below (also see Figure \ref{fig:ordering-strategies} for illustration):

\begin{small}
\begin{enumerate}
    \item \textsc{total dependency length}: Summation of head-dependent distances in a dependency tree of a sentence. And the head-dependent distance was estimated by counting the number of intervening words between them.
    \item \textsc{constituent dependency length}: Number of words spanned between the head of the preverbal constituent and the main root on which the constituent is dependent.
    \item \textsc{constituent length}: Total word counts in the constituent.
\end{enumerate}
\end{small}

\subsection{Regression analysis}

In this section, we test our main hypothesis that speakers optimize word order choices in Hindi by adopting least-effort strategy (i.e., shifting the shortest constituent closest to the main verb) under the constraints of dependency length minimization. Figure \ref{var-const-counts} depicts the distribution of variants with varying preverbal constituents. Since our classification dataset is dominated by reference-variant pairs with 5 preverbal constituents, we test the least-effort hypothesis on the transformed version of our dataset with 5 preverbal constituents only (87143 data points) using a logistic regression model. We estimated the dependency length and constituent length of all 5 preverbal constituents and introduced them as predictors in our regression models ($choice~\sim~\delta const1+\delta const2+ \delta const3+ \delta const4+ \delta const5$). However, before doing so, we first deployed the Recursive Feature Elimination Cross-validation algorithm~\cite[RFECV]{guyon2002gene} to find the best predictors among five preverbal constituents and then reported the regression results using the obtained optimal features. 

Tables \ref{tab:deplen-reg-only5} and \ref{tab:constlen-reg-only5} display the results of our regression analysis. Negative regression coefficients in Tables \ref{tab:deplen-reg-only5} and \ref{tab:constlen-reg-only5} suggest that reference sentences tend to have lower constituent-level dependency length and the length of constituent, respectively, than their paired variant.  The larger magnitude of the regression coefficient of the last two preverbal constituents closest to the verb ($const4~\&~const5$) as opposed to other preverbal constituents in the model suggests their greater role in minimizing overall dependency length of reference sentence than the paired variants. Notably, our results were consistent across reference-variant sentences with 2 to 6 preverbal constituents, indicating that our findings are not limited to rare constructions, but are applicable to frequent ones as well in the Hindi natural corpus. In sum, these findings suggest that speakers indeed, apply the least-effort strategy to optimize for word orders such that the preverbal constituents closest to the verb are shorter and also have lower dependency length.

\begin{table}[t]
\centering
\scalebox{0.95}{
\begin{tabular}{|l|c|c|c|}
\hline
\textbf{Predictor} & \textbf{Estimate} & \textbf{Std-Error} & \textbf{z-value} \\ \hline
Intercept & \textbf{0.018} & 0.008 & 1.98 \\
$1^{st}$ constituent's deplen & {-0.004} & 0.012 & -0.35 \\
$2^{nd}$ constituent's deplen & \textbf{0.088} & 0.014 & 6.20 \\
$3^{rd}$ constituent's deplen & \textbf{-0.147} & 0.015 & -10.12 \\
$4^{th}$ constituent's deplen & \textbf{-0.529} & 0.016 & -32.90 \\
$5^{th}$ constituent's deplen & \textbf{-2.720} & 0.023 & -118.03 \\\hline
\end{tabular}}
\caption{Regression model containing dependency lengths (deplen) of preverbal constituents as predictors (87143 data points); significant predictors denoted in bold with p $<$ 0.001}
\label{tab:deplen-reg-only5}
\end{table}

\begin{table}[t]
\centering
\scalebox{0.95}{
\begin{tabular}{|l|c|c|c|}
\hline
\textbf{Predictor} & \textbf{Estimate} & \textbf{Std-Error} & \textbf{z-value} \\ \hline
Intercept & -0.003 & 0.008 & -0.41  \\
$1^{st}$ constituent's length & \textbf{-0.083} & 0.009 & -8.44 \\
$3^{rd}$ constituent's length & \textbf{0.058} & 0.010 & 5.72 \\
$4^{th}$ constituent's length & \textbf{-0.148} & 0.009 & -15.26 \\
$5^{th}$ constituent's length & \textbf{-1.549} & 0.016 & -97.82  \\\hline
\end{tabular}}
\caption{Regression model containing preverbal constituent lengths as predictors (87143 data points); significant predictors denoted in bold with p $<$ 0.001; constituent length = word counts in the constituent}
\label{tab:constlen-reg-only5}
\end{table}


\begin{table}[t]
\centering
\scalebox{0.95}{
\begin{tabular}{l|c}
\hline 
Predictor(s) & Accuracy \\ \hline 
total dependency length & 62.69 \\
2nd-last preverbal constituent's deplen & 68.48*** \\
last preverbal constituent's deplen & 72.70*** \\
\hline
last + 2nd last preverbal constituent's deplen & 77.17*** \\
\hline 
\end{tabular}}
\caption{Prediction accuracy of distinct models with dependency length as predictor on full dataset (184818 data points; deplen = dependency length; McNemar's two-tailed significance test compared to previous row: *** $p<0.001$)}
\label{tab:pred-results-dl}
\end{table}


\subsection{Classification analysis}

Based on the insights from the previous section, we deployed the last two preverbal constituent's dependency length and constituent length (closest to the main verb: \textit{2nd~last}~\&~\textit{last~constituents}) as predictors into the classification model aimed at predicting reference sentences amidst the counterfactual variants. We make use of our entire dataset (184818 reference-variant pairs) and conduct 10-fold cross-validation to evaluate model's classification accuracy, \emph{i.e.,} the percentage of data points where a model correctly predicted the referent sentence over a paired variant. Tables \ref{tab:pred-results-dl} and \ref{tab:pred-results-cl} present the prediction performance of our models. As illustrated in Table \ref{tab:pred-results-dl}, in terms of individual classification accuracy, the dependency length of the last constituent turned out to be the best predictor (72.70\%), thus validating the efficacy of the proposed least-effort principle. Interestingly, this measure supersedes the performance of the overall dependency length measure (62.69\%) of the sentence as well.

Over a baseline model containing the last constituent's dependency length, adding the 2nd last constituent's dependency length induced a significant increase of 4.47\% in the classification accuracy (p $<$ 0.001 using McNemar's two-tailed test). We find similar results when the constituent lengths of the last two constituents were deployed as predictors in the classifier, as shown in Table \ref{tab:pred-results-cl}. These findings lend credence to the idea that speakers adopt least-effort strategy, as measured by dependency length minimization, in determining Hindi word-order preferences.

\begin{table}[t]
\centering
\scalebox{0.95}{
\begin{tabular}{l|c}
\hline 
Predictor(s) & Accuracy \\ \hline 
2nd-last preverbal constituent length & 54.35 \\
last preverbal constituent length & 69.62*** \\\hline
last + 2nd last preverbal constituent length & 70.28*** \\
\hline 
\end{tabular}}
\caption{Prediction accuracy of distinct models with constituent length as predictor (184818 data points; McNemar's two-tailed test compared to previous row: *** $p<0.001$)}
\label{tab:pred-results-cl}
\end{table}


\section{Discussion and conclusion}\label{sect:disc}

Our main findings demonstrate that speakers in Hindi choose optimal word orders by placing the shortest phrase closest to the main verb, consistent with the central ideas of the bounded rationality view of decision making.  This least effort approach not only reduces the overall dependency length but also enables a selective search procedure in a vast search space where finding the optimal solution is difficult. Therefore the sentence planning system uses a simple yet powerful heuristic.  Additionally, as depicted in Figure \ref{fig:preverbal-constlen}, the pressure to shorten the constituent closest to the main verb increases as the length of the sentence increases. We also observed that sentences that originally appeared in the corpus had dependency lengths distribution similar to the proposed least-effort solution evincing the efficacy of bounded rationality in choosing the appropriate word order in Hindi (See Figure \ref{fig:bounded-rationality}). In addition, the HUTB corpus sentences had lower overall dependency length than the random word orders consistent with the earlier studies~\cite{Liu2008,futrell2015,futrell2020dependency}. 

The prediction performance of two distinct classification models aimed at distinguishing reference sentences amidst artificially generated variants using just the constituent lengths (70.28\%) and just the dependency lengths of the last two preverbal constituents that are nearest to the main verb (77.17\%) emphasizes the significance of the proposed least-effort strategy in shaping the constituent order~\cite{Liu2017,temperley-gildea-ar18,futrell2020dependency}. It's also interesting to note that the overall dependency length (62.69\%) performs poorly in predicting corpus reference sentences (amidst variants) than the least-effort strategy (last preverbal constituent's dependency length = 72.70\%; last preverbal constituent length = 69.62\%). As a part of future work, we plan to examine the effectiveness of our proposed least-effort measure across various languages from different language families. Studying how languages may differ in the way they implement bounded rationality strategies can provide further insight into the relationship between language and cognition.

The broader implication of our work is that word order choices can be seen as a result of the pressures for efficient communication within the constraints of bounded rationality~\cite{gigerenzer1996reasoning,gigerenzer2015simply}. The speaker's attempt to minimize dependency length is one such frugal strategy to deal with limited cognitive resources and the complexity of the task. Given that we often lack the cognitive and computational resources to solve problems exactly, we tend to use approximation methods as a more viable strategy, leading to bounded rationality~\cite{simon1990invariants,gigerenzer2002bounded}. According to \citeauthor{simon1990invariants}, the following mechanisms are used by people to deal with complex problems and make decisions under the constraints of bounded rationality -- \textit{pattern recognition, heuristic search,} and \textit{extrapolation}. Along similar lines, in the case of language, communicative efficiency can be achieved by using most \textit{common word-order patterns}, doing \textit{selective search} among possible word orders that place the syntactically related words close together or salient/predictable words early in the sentence to overcome the memory load. In addition, speakers can make use to \textit{certain properties of languages}, such as word-order flexibility, which provides helpful cues to the listener for improved understanding. From the listener's perspective, flexibility in word order can be challenging, and relying on various language-specific cues could be a rational strategy for them to navigate this challenge within the constraints of bounded rationality. Therefore, our results and interpretations of bounded rationality hold from both speaker as well as hearer perspectives. A thorough investigation from these perspectives will pave the way for better understanding.  To make more tangible claims about language production in general, it will also be necessary to investigate the efficacy of the least-effort principle on spoken data.

Although bounded rationality may not fully account for all aspects of language, it may provide an explanation for certain aspects, such as the extent of dependency length minimization, word-order patterns, individual differences in decision-making processes, and the cognitive mechanisms involved in language use. Future work needs to investigate how different factors, such as context, task demands, and individual cognitive differences, may affect how speakers and listeners use boundedly rational strategies.

Overall, our results suggest that within the constraints of bounded rationality, dependency length minimization is a significant predictor of Hindi word order patterns.

\section{Acknowledgments}

We would like to thank the anonymous CogSci-2023 reviewers for their helpful suggestions and feedback.

\bibliographystyle{apacite}

\setlength{\bibleftmargin}{.125in}
\setlength{\bibindent}{-\bibleftmargin}

\bibliography{CogSci_Template,bibfile,extra}

\begin{thebibliography}{}

\bibitem [\protect \citeauthoryear {%
Arnold%
, Wasow%
, Losongco%
\BCBL {}\ \BBA {} Ginstrom%
}{%
Arnold%
\ \protect \BOthers {.}}{%
{\protect \APACyear {2000}}%
}]{%
Arnold00heaviness}
\APACinsertmetastar {%
Arnold00heaviness}%
\begin{APACrefauthors}%
Arnold, J\BPBI E.%
, Wasow, T.%
, Losongco, A.%
\BCBL {}\ \BBA {} Ginstrom, R.%
\end{APACrefauthors}%
\unskip\
\newblock
\APACrefYearMonthDay{2000}{}{}.
\newblock
{\BBOQ}\APACrefatitle {Heaviness vs.\ Newness: The Effects of Structural
  Complexity and Discourse Status on Constituent Ordering} {Heaviness vs.\
  newness: The effects of structural complexity and discourse status on
  constituent ordering}.{\BBCQ}
\newblock
\APACjournalVolNumPages{Language}{76}{}{28--55}.
\newblock
\begin{APACrefURL} \url{https://www.jstor.org/stable/417392} \end{APACrefURL}
\PrintBackRefs{\CurrentBib}

\bibitem [\protect \citeauthoryear {%
Bhatt%
\ \protect \BOthers {.}}{%
Bhatt%
\ \protect \BOthers {.}}{%
{\protect \APACyear {2009}}%
}]{%
Bhatt2009}
\APACinsertmetastar {%
Bhatt2009}%
\begin{APACrefauthors}%
Bhatt, R.%
, Narasimhan, B.%
, Palmer, M.%
, Rambow, O.%
, Sharma, D\BPBI M.%
\BCBL {}\ \BBA {} Xia, F.%
\end{APACrefauthors}%
\unskip\
\newblock
\APACrefYearMonthDay{2009}{}{}.
\newblock
{\BBOQ}\APACrefatitle {A Multi-representational and Multi-layered Treebank for
  {H}indi/Urdu} {A multi-representational and multi-layered treebank for
  {H}indi/urdu}.{\BBCQ}
\newblock
\BIn{} \APACrefbtitle {Proceedings of the Third Linguistic Annotation Workshop}
  {Proceedings of the third linguistic annotation workshop}\ (\BPGS\ 186--189).
\newblock
\APACaddressPublisher{Stroudsburg, PA, USA}{Association for Computational
  Linguistics}.
\newblock
\begin{APACrefURL} \url{http://dl.acm.org/citation.cfm?id=1698381.1698417}
  \end{APACrefURL}
\PrintBackRefs{\CurrentBib}

\bibitem [\protect \citeauthoryear {%
Bock%
\ \BBA {} Warren%
}{%
Bock%
\ \BBA {} Warren%
}{%
{\protect \APACyear {1985}}%
}]{%
Bock1985}
\APACinsertmetastar {%
Bock1985}%
\begin{APACrefauthors}%
Bock, J\BPBI K.%
\BCBT {}\ \BBA {} Warren, R\BPBI K.%
\end{APACrefauthors}%
\unskip\
\newblock
\APACrefYearMonthDay{1985}{}{}.
\newblock
{\BBOQ}\APACrefatitle {{Conceptual accessibility and syntactic structure in
  sentence formulation}} {{Conceptual accessibility and syntactic structure in
  sentence formulation}}.{\BBCQ}
\newblock
\APACjournalVolNumPages{Cognition}{21}{}{47--67}.
\newblock
\begin{APACrefURL}
  \url{https://www.sciencedirect.com/science/article/pii/001002778590023X}
  \end{APACrefURL}
\PrintBackRefs{\CurrentBib}

\bibitem [\protect \citeauthoryear {%
Choi%
}{%
Choi%
}{%
{\protect \APACyear {2007}}%
}]{%
Choi2007}
\APACinsertmetastar {%
Choi2007}%
\begin{APACrefauthors}%
Choi, H\BHBI w.%
\end{APACrefauthors}%
\unskip\
\newblock
\APACrefYearMonthDay{2007}{}{}.
\newblock
{\BBOQ}\APACrefatitle {{Length and Order: A Corpus Study of Korean
  Dative-Accusative Construction}} {{Length and Order: A Corpus Study of Korean
  Dative-Accusative Construction}}.{\BBCQ}
\newblock
\APACjournalVolNumPages{Discourse and Cognition}{14}{3}{207--227}.
\newblock
\begin{APACrefURL}
  \url{https://www.dbpia.co.kr/journal/articleDetail?nodeId=NODE00986720}
  \end{APACrefURL}
\PrintBackRefs{\CurrentBib}

\bibitem [\protect \citeauthoryear {%
Faghiri%
\ \BBA {} Samvelian%
}{%
Faghiri%
\ \BBA {} Samvelian%
}{%
{\protect \APACyear {2020}}%
}]{%
faghiri2020word}
\APACinsertmetastar {%
faghiri2020word}%
\begin{APACrefauthors}%
Faghiri, P.%
\BCBT {}\ \BBA {} Samvelian, P.%
\end{APACrefauthors}%
\unskip\
\newblock
\APACrefYearMonthDay{2020}{}{}.
\newblock
{\BBOQ}\APACrefatitle {Word order preferences and the effect of phrasal length
  in SOV languages: evidence from sentence production in Persian} {Word order
  preferences and the effect of phrasal length in sov languages: evidence from
  sentence production in persian}.{\BBCQ}
\newblock
\APACjournalVolNumPages{Glossa: a journal of general linguistics
  (2016-2021)}{5}{1}{86}.
\PrintBackRefs{\CurrentBib}

\bibitem [\protect \citeauthoryear {%
{Ferrer-i-Cancho}%
}{%
{Ferrer-i-Cancho}%
}{%
{\protect \APACyear {2004}}%
}]{%
i2004euclidean}
\APACinsertmetastar {%
i2004euclidean}%
\begin{APACrefauthors}%
{Ferrer-i-Cancho}, R.%
\end{APACrefauthors}%
\unskip\
\newblock
\APACrefYearMonthDay{2004}{}{}.
\newblock
{\BBOQ}\APACrefatitle {Euclidean distance between syntactically linked words}
  {Euclidean distance between syntactically linked words}.{\BBCQ}
\newblock
\APACjournalVolNumPages{Physical Review E}{70}{5}{056135}.
\newblock
\begin{APACrefURL} \url{https://core.ac.uk/download/pdf/286456939.pdf}
  \end{APACrefURL}
\PrintBackRefs{\CurrentBib}

\bibitem [\protect \citeauthoryear {%
Futrell%
, Levy%
\BCBL {}\ \BBA {} Gibson%
}{%
Futrell%
\ \protect \BOthers {.}}{%
{\protect \APACyear {2020}}%
}]{%
futrell2020dependency}
\APACinsertmetastar {%
futrell2020dependency}%
\begin{APACrefauthors}%
Futrell, R.%
, Levy, R\BPBI P.%
\BCBL {}\ \BBA {} Gibson, E.%
\end{APACrefauthors}%
\unskip\
\newblock
\APACrefYearMonthDay{2020}{}{}.
\newblock
{\BBOQ}\APACrefatitle {Dependency locality as an explanatory principle for word
  order} {Dependency locality as an explanatory principle for word
  order}.{\BBCQ}
\newblock
\APACjournalVolNumPages{Language}{96}{2}{371--412}.
\newblock
\begin{APACrefURL}
  \url{https://sites.socsci.uci.edu/~rfutrell/papers/futrell2020dependency.pdf}
  \end{APACrefURL}
\PrintBackRefs{\CurrentBib}

\bibitem [\protect \citeauthoryear {%
Futrell%
, Mahowald%
\BCBL {}\ \BBA {} Gibson%
}{%
Futrell%
\ \protect \BOthers {.}}{%
{\protect \APACyear {2015}}%
}]{%
futrell2015}
\APACinsertmetastar {%
futrell2015}%
\begin{APACrefauthors}%
Futrell, R.%
, Mahowald, K.%
\BCBL {}\ \BBA {} Gibson, E.%
\end{APACrefauthors}%
\unskip\
\newblock
\APACrefYearMonthDay{2015}{}{}.
\newblock
{\BBOQ}\APACrefatitle {Large-scale evidence of dependency length minimization
  in 37 languages} {Large-scale evidence of dependency length minimization in
  37 languages}.{\BBCQ}
\newblock
\APACjournalVolNumPages{Proceedings of the National Academy of
  Sciences}{112}{33}{10336-10341}.
\newblock
\begin{APACrefURL} \url{http://www.pnas.org/content/112/33/10336.abstract}
  \end{APACrefURL}
\newblock
\begin{APACrefDOI} \doi{10.1073/pnas.1502134112} \end{APACrefDOI}
\PrintBackRefs{\CurrentBib}

\bibitem [\protect \citeauthoryear {%
Gibson%
}{%
Gibson%
}{%
{\protect \APACyear {1991}}%
}]{%
gibson1991computational}
\APACinsertmetastar {%
gibson1991computational}%
\begin{APACrefauthors}%
Gibson, E.%
\end{APACrefauthors}%
\unskip\
\newblock
\APACrefYear{1991}.
\newblock
\APACrefbtitle {A computational theory of human linguistic processing: Memory
  limitations and processing breakdown} {A computational theory of human
  linguistic processing: Memory limitations and processing breakdown}.
\newblock
\APACaddressPublisher{}{Carnegie Mellon University}.
\newblock
\begin{APACrefURL}
  \url{http://tedlab.mit.edu/tedlab_website/researchpapers/Gibson_1991_PhDthesis.pdf}
  \end{APACrefURL}
\PrintBackRefs{\CurrentBib}

\bibitem [\protect \citeauthoryear {%
Gibson%
}{%
Gibson%
}{%
{\protect \APACyear {1998}}%
}]{%
Gib98}
\APACinsertmetastar {%
Gib98}%
\begin{APACrefauthors}%
Gibson, E.%
\end{APACrefauthors}%
\unskip\
\newblock
\APACrefYearMonthDay{1998}{}{}.
\newblock
{\BBOQ}\APACrefatitle {Linguistic complexity: Locality of syntactic
  dependencies} {Linguistic complexity: Locality of syntactic
  dependencies}.{\BBCQ}
\newblock
\APACjournalVolNumPages{Cognition}{68}{}{1--76}.
\newblock
\begin{APACrefURL} \url{https://doi.org/10.1016/S0010-0277(98)00034-1}
  \end{APACrefURL}
\PrintBackRefs{\CurrentBib}

\bibitem [\protect \citeauthoryear {%
Gibson%
}{%
Gibson%
}{%
{\protect \APACyear {2000}}%
}]{%
gibson00}
\APACinsertmetastar {%
gibson00}%
\begin{APACrefauthors}%
Gibson, E.%
\end{APACrefauthors}%
\unskip\
\newblock
\APACrefYearMonthDay{2000}{}{}.
\newblock
{\BBOQ}\APACrefatitle {Dependency Locality Theory: {A} Distance-Based Theory of
  Linguistic Complexity} {Dependency locality theory: {A} distance-based theory
  of linguistic complexity}.{\BBCQ}
\newblock
\BIn{} A.~Marantz, Y.~Miyashita\BCBL {}\ \BBA {} W.~O'Neil\ (\BEDS),
  \APACrefbtitle {Image, Language, brain: {P}apers from the First Mind
  Articulation Project Symposium.} {Image, language, brain: {P}apers from the
  first mind articulation project symposium.}
\newblock
\APACaddressPublisher{Cambridge, MA}{MIT Press}.
\newblock
\begin{APACrefURL}
  \url{http://www.ling.uni-potsdam.de/~vasishth/Papers/Gibson-Cognition2000.pdf}
  \end{APACrefURL}
\PrintBackRefs{\CurrentBib}

\bibitem [\protect \citeauthoryear {%
Gibson%
\ \protect \BOthers {.}}{%
Gibson%
\ \protect \BOthers {.}}{%
{\protect \APACyear {2019}}%
}]{%
Gibson2019}
\APACinsertmetastar {%
Gibson2019}%
\begin{APACrefauthors}%
Gibson, E.%
, Futrell, R.%
, Piandadosi, S\BPBI T.%
, Dautriche, I.%
, Mahowald, K.%
, Bergen, L.%
\BCBL {}\ \BBA {} Levy, R.%
\end{APACrefauthors}%
\unskip\
\newblock
\APACrefYearMonthDay{2019}{May}{01}.
\newblock
{\BBOQ}\APACrefatitle {How Efficiency Shapes Human Language} {How efficiency
  shapes human language}.{\BBCQ}
\newblock
\APACjournalVolNumPages{Trends in Cognitive Sciences}{23}{5}{389-407}.
\newblock
\begin{APACrefURL} \url{https://doi.org/10.1016/j.tics.2019.02.003}
  \end{APACrefURL}
\newblock
\begin{APACrefDOI} \doi{10.1016/j.tics.2019.02.003} \end{APACrefDOI}
\PrintBackRefs{\CurrentBib}

\bibitem [\protect \citeauthoryear {%
Gigerenzer%
}{%
Gigerenzer%
}{%
{\protect \APACyear {2015}}%
}]{%
gigerenzer2015simply}
\APACinsertmetastar {%
gigerenzer2015simply}%
\begin{APACrefauthors}%
Gigerenzer, G.%
\end{APACrefauthors}%
\unskip\
\newblock
\APACrefYear{2015}.
\newblock
\APACrefbtitle {Simply rational: Decision making in the real world} {Simply
  rational: Decision making in the real world}.
\newblock
\APACaddressPublisher{}{Evolution and Cognition}.
\PrintBackRefs{\CurrentBib}

\bibitem [\protect \citeauthoryear {%
Gigerenzer%
\ \BBA {} Gaissmaier%
}{%
Gigerenzer%
\ \BBA {} Gaissmaier%
}{%
{\protect \APACyear {2011}}%
}]{%
gigerenzer2011heuristic}
\APACinsertmetastar {%
gigerenzer2011heuristic}%
\begin{APACrefauthors}%
Gigerenzer, G.%
\BCBT {}\ \BBA {} Gaissmaier, W.%
\end{APACrefauthors}%
\unskip\
\newblock
\APACrefYearMonthDay{2011}{}{}.
\newblock
{\BBOQ}\APACrefatitle {Heuristic decision making} {Heuristic decision
  making}.{\BBCQ}
\newblock
\APACjournalVolNumPages{Annual review of psychology}{62}{}{451--482}.
\PrintBackRefs{\CurrentBib}

\bibitem [\protect \citeauthoryear {%
Gigerenzer%
\ \BBA {} Goldstein%
}{%
Gigerenzer%
\ \BBA {} Goldstein%
}{%
{\protect \APACyear {1996}}%
}]{%
gigerenzer1996reasoning}
\APACinsertmetastar {%
gigerenzer1996reasoning}%
\begin{APACrefauthors}%
Gigerenzer, G.%
\BCBT {}\ \BBA {} Goldstein, D\BPBI G.%
\end{APACrefauthors}%
\unskip\
\newblock
\APACrefYearMonthDay{1996}{}{}.
\newblock
{\BBOQ}\APACrefatitle {Reasoning the fast and frugal way: models of bounded
  rationality.} {Reasoning the fast and frugal way: models of bounded
  rationality.}{\BBCQ}
\newblock
\APACjournalVolNumPages{Psychological review}{103}{4}{650}.
\PrintBackRefs{\CurrentBib}

\bibitem [\protect \citeauthoryear {%
Gigerenzer%
, Hertwig%
\BCBL {}\ \BBA {} Pachur%
}{%
Gigerenzer%
\ \protect \BOthers {.}}{%
{\protect \APACyear {2011}}%
}]{%
gigerenzer2011heuristicsbook}
\APACinsertmetastar {%
gigerenzer2011heuristicsbook}%
\begin{APACrefauthors}%
Gigerenzer, G.%
, Hertwig, R\BPBI E.%
\BCBL {}\ \BBA {} Pachur, T\BPBI E.%
\end{APACrefauthors}%
\unskip\
\newblock
\APACrefYear{2011}.
\newblock
\APACrefbtitle {Heuristics: The foundations of adaptive behavior.} {Heuristics:
  The foundations of adaptive behavior.}
\newblock
\APACaddressPublisher{}{Oxford University Press}.
\PrintBackRefs{\CurrentBib}

\bibitem [\protect \citeauthoryear {%
Gigerenzer%
\ \BBA {} Selten%
}{%
Gigerenzer%
\ \BBA {} Selten%
}{%
{\protect \APACyear {2002}}%
}]{%
gigerenzer2002bounded}
\APACinsertmetastar {%
gigerenzer2002bounded}%
\begin{APACrefauthors}%
Gigerenzer, G.%
\BCBT {}\ \BBA {} Selten, R.%
\end{APACrefauthors}%
\unskip\
\newblock
\APACrefYear{2002}.
\newblock
\APACrefbtitle {Bounded rationality: The adaptive toolbox} {Bounded
  rationality: The adaptive toolbox}.
\newblock
\APACaddressPublisher{}{MIT press}.
\PrintBackRefs{\CurrentBib}

\bibitem [\protect \citeauthoryear {%
Gildea%
\ \BBA {} Temperley%
}{%
Gildea%
\ \BBA {} Temperley%
}{%
{\protect \APACyear {2010}}%
}]{%
GildeaT10}
\APACinsertmetastar {%
GildeaT10}%
\begin{APACrefauthors}%
Gildea, D.%
\BCBT {}\ \BBA {} Temperley, D.%
\end{APACrefauthors}%
\unskip\
\newblock
\APACrefYearMonthDay{2010}{}{}.
\newblock
{\BBOQ}\APACrefatitle {Do Grammars Minimize Dependency Length?} {Do grammars
  minimize dependency length?}{\BBCQ}
\newblock
\APACjournalVolNumPages{Cognitive Science}{34}{2}{286--310}.
\PrintBackRefs{\CurrentBib}

\bibitem [\protect \citeauthoryear {%
Guyon%
, Weston%
, Barnhill%
\BCBL {}\ \BBA {} Vapnik%
}{%
Guyon%
\ \protect \BOthers {.}}{%
{\protect \APACyear {2002}}%
}]{%
guyon2002gene}
\APACinsertmetastar {%
guyon2002gene}%
\begin{APACrefauthors}%
Guyon, I.%
, Weston, J.%
, Barnhill, S.%
\BCBL {}\ \BBA {} Vapnik, V.%
\end{APACrefauthors}%
\unskip\
\newblock
\APACrefYearMonthDay{2002}{}{}.
\newblock
{\BBOQ}\APACrefatitle {Gene selection for cancer classification using support
  vector machines} {Gene selection for cancer classification using support
  vector machines}.{\BBCQ}
\newblock
\APACjournalVolNumPages{Machine learning}{46}{}{389--422}.
\newblock
\begin{APACrefURL}
  \url{https://link.springer.com/article/10.1023/A:1012487302797}
  \end{APACrefURL}
\PrintBackRefs{\CurrentBib}

\bibitem [\protect \citeauthoryear {%
Hawkins%
}{%
Hawkins%
}{%
{\protect \APACyear {1994}}%
}]{%
hawkins1994}
\APACinsertmetastar {%
hawkins1994}%
\begin{APACrefauthors}%
Hawkins, J\BPBI A.%
\end{APACrefauthors}%
\unskip\
\newblock
\APACrefYear{1994}.
\newblock
\APACrefbtitle {A {P}erformance Theory of Order and Constituency} {A
  {P}erformance theory of order and constituency}.
\newblock
\APACaddressPublisher{New York}{Cambridge University Press}.
\PrintBackRefs{\CurrentBib}

\bibitem [\protect \citeauthoryear {%
Hawkins%
}{%
Hawkins%
}{%
{\protect \APACyear {2004}}%
}]{%
hawkins04}
\APACinsertmetastar {%
hawkins04}%
\begin{APACrefauthors}%
Hawkins, J\BPBI A.%
\end{APACrefauthors}%
\unskip\
\newblock
\APACrefYear{2004}.
\newblock
\APACrefbtitle {Efficiency and Complexity in Grammars} {Efficiency and
  complexity in grammars}.
\newblock
\APACaddressPublisher{}{Oxford University Press}.
\PrintBackRefs{\CurrentBib}

\bibitem [\protect \citeauthoryear {%
Hudson%
}{%
Hudson%
}{%
{\protect \APACyear {1995}}%
}]{%
hudson1995measuring}
\APACinsertmetastar {%
hudson1995measuring}%
\begin{APACrefauthors}%
Hudson, R\BPBI A.%
\end{APACrefauthors}%
\unskip\
\newblock
\APACrefYearMonthDay{1995}{}{}.
\newblock
{\BBOQ}\APACrefatitle {Measuring Syntactic Difficulty} {Measuring syntactic
  difficulty}.{\BBCQ}
\newblock
\APACjournalVolNumPages{University College London}{}{}{}.
\newblock
\begin{APACrefURL}
  \url{http://dickhudson.com/wp-content/uploads/2013/07/Difficulty.pdf}
  \end{APACrefURL}
\PrintBackRefs{\CurrentBib}

\bibitem [\protect \citeauthoryear {%
Jaeger%
\ \BBA {} Norcliffe%
}{%
Jaeger%
\ \BBA {} Norcliffe%
}{%
{\protect \APACyear {2009}}%
}]{%
Jaeger2009compass}
\APACinsertmetastar {%
Jaeger2009compass}%
\begin{APACrefauthors}%
Jaeger, T\BPBI F.%
\BCBT {}\ \BBA {} Norcliffe, E.%
\end{APACrefauthors}%
\unskip\
\newblock
\APACrefYearMonthDay{2009}{}{}.
\newblock
{\BBOQ}\APACrefatitle {The cross-linguistic study of sentence production: State
  of the art and a call for action} {The cross-linguistic study of sentence
  production: State of the art and a call for action}.{\BBCQ}
\newblock
\APACjournalVolNumPages{Language and Linguistic Compass}{3}{4}{866--887}.
\newblock
\begin{APACrefURL} \url{http://dx.doi.org/10.1111/j.1749-818X.2009.00147.x}
  \end{APACrefURL}
\PrintBackRefs{\CurrentBib}

\bibitem [\protect \citeauthoryear {%
Joachims%
}{%
Joachims%
}{%
{\protect \APACyear {2002}}%
}]{%
Joachims:2002}
\APACinsertmetastar {%
Joachims:2002}%
\begin{APACrefauthors}%
Joachims, T.%
\end{APACrefauthors}%
\unskip\
\newblock
\APACrefYearMonthDay{2002}{}{}.
\newblock
{\BBOQ}\APACrefatitle {Optimizing Search Engines Using Clickthrough Data}
  {Optimizing search engines using clickthrough data}.{\BBCQ}
\newblock
\BIn{} \APACrefbtitle {Proceedings of the Eighth ACM SIGKDD International
  Conference on Knowledge Discovery and Data Mining} {Proceedings of the eighth
  acm sigkdd international conference on knowledge discovery and data mining}\
  (\BPGS\ 133--142).
\newblock
\APACaddressPublisher{New York, NY, USA}{ACM}.
\newblock
\begin{APACrefURL} \url{http://doi.acm.org/10.1145/775047.775067}
  \end{APACrefURL}
\newblock
\begin{APACrefDOI} \doi{10.1145/775047.775067} \end{APACrefDOI}
\PrintBackRefs{\CurrentBib}

\bibitem [\protect \citeauthoryear {%
Kachru%
}{%
Kachru%
}{%
{\protect \APACyear {1982}}%
}]{%
kachru1982}
\APACinsertmetastar {%
kachru1982}%
\begin{APACrefauthors}%
Kachru, Y.%
\end{APACrefauthors}%
\unskip\
\newblock
\APACrefYearMonthDay{1982}{}{}.
\newblock
{\BBOQ}\APACrefatitle {Conjunct Verbs in Hindi-Urdu and Persian} {Conjunct
  verbs in hindi-urdu and persian}.{\BBCQ}
\newblock
\APACjournalVolNumPages{South Asian Review}{6}{3}{117-126}.
\newblock
\begin{APACrefURL} \url{https://doi.org/10.1080/02759527.1982.11933096}
  \end{APACrefURL}
\newblock
\begin{APACrefDOI} \doi{10.1080/02759527.1982.11933096} \end{APACrefDOI}
\PrintBackRefs{\CurrentBib}

\bibitem [\protect \citeauthoryear {%
Liu%
}{%
Liu%
}{%
{\protect \APACyear {2008}}%
}]{%
Liu2008}
\APACinsertmetastar {%
Liu2008}%
\begin{APACrefauthors}%
Liu, H.%
\end{APACrefauthors}%
\unskip\
\newblock
\APACrefYearMonthDay{2008}{}{}.
\newblock
{\BBOQ}\APACrefatitle {Dependency distance as a metric of language
  comprehension difficulty} {Dependency distance as a metric of language
  comprehension difficulty}.{\BBCQ}
\newblock
\APACjournalVolNumPages{Journal of Cognitive Science}{9}{2}{159-191}.
\newblock
\begin{APACrefURL} \url{http://www.lingviko.net/JCS.pdf} \end{APACrefURL}
\PrintBackRefs{\CurrentBib}

\bibitem [\protect \citeauthoryear {%
Liu%
, Xu%
\BCBL {}\ \BBA {} Liang%
}{%
Liu%
\ \protect \BOthers {.}}{%
{\protect \APACyear {2017}}%
}]{%
Liu2017}
\APACinsertmetastar {%
Liu2017}%
\begin{APACrefauthors}%
Liu, H.%
, Xu, C.%
\BCBL {}\ \BBA {} Liang, J.%
\end{APACrefauthors}%
\unskip\
\newblock
\APACrefYearMonthDay{2017}{}{}.
\newblock
{\BBOQ}\APACrefatitle {Dependency distance: A new perspective on syntactic
  patterns in natural languages} {Dependency distance: A new perspective on
  syntactic patterns in natural languages}.{\BBCQ}
\newblock
\APACjournalVolNumPages{Physics of Life Reviews}{21}{}{171 - 193}.
\newblock
\begin{APACrefURL}
  \url{http://www.sciencedirect.com/science/article/pii/S1571064517300532}
  \end{APACrefURL}
\newblock
\begin{APACrefDOI} \doi{https://doi.org/10.1016/j.plrev.2017.03.002}
  \end{APACrefDOI}
\PrintBackRefs{\CurrentBib}

\bibitem [\protect \citeauthoryear {%
Newell%
\ \BBA {} Simon%
}{%
Newell%
\ \BBA {} Simon%
}{%
{\protect \APACyear {1972}}%
}]{%
NewellSimon1972a}
\APACinsertmetastar {%
NewellSimon1972a}%
\begin{APACrefauthors}%
Newell, A.%
\BCBT {}\ \BBA {} Simon, H\BPBI A.%
\end{APACrefauthors}%
\unskip\
\newblock
\APACrefYear{1972}.
\newblock
\APACrefbtitle {Human problem solving} {Human problem solving}.
\newblock
\APACaddressPublisher{Englewood Cliffs, NJ}{Prentice-Hall}.
\PrintBackRefs{\CurrentBib}

\bibitem [\protect \citeauthoryear {%
Ranjan%
, Rajkumar%
\BCBL {}\ \BBA {} Agarwal%
}{%
Ranjan%
\ \protect \BOthers {.}}{%
{\protect \APACyear {2022}}%
}]{%
cog:sid}
\APACinsertmetastar {%
cog:sid}%
\begin{APACrefauthors}%
Ranjan, S.%
, Rajkumar, R.%
\BCBL {}\ \BBA {} Agarwal, S.%
\end{APACrefauthors}%
\unskip\
\newblock
\APACrefYearMonthDay{2022}{}{}.
\newblock
{\BBOQ}\APACrefatitle {Locality and expectation effects in Hindi preverbal
  constituent ordering} {Locality and expectation effects in hindi preverbal
  constituent ordering}.{\BBCQ}
\newblock
\APACjournalVolNumPages{Cognition}{223}{}{104959}.
\newblock
\begin{APACrefURL}
  \url{https://www.sciencedirect.com/science/article/pii/S0010027721003826}
  \end{APACrefURL}
\newblock
\begin{APACrefDOI} \doi{https://doi.org/10.1016/j.cognition.2021.104959}
  \end{APACrefDOI}
\PrintBackRefs{\CurrentBib}

\bibitem [\protect \citeauthoryear {%
Sharma%
, Futrell%
\BCBL {}\ \BBA {} Husain%
}{%
Sharma%
\ \protect \BOthers {.}}{%
{\protect \APACyear {2020}}%
}]{%
sharma2020determines}
\APACinsertmetastar {%
sharma2020determines}%
\begin{APACrefauthors}%
Sharma, K.%
, Futrell, R.%
\BCBL {}\ \BBA {} Husain, S.%
\end{APACrefauthors}%
\unskip\
\newblock
\APACrefYearMonthDay{2020}{}{}.
\newblock
{\BBOQ}\APACrefatitle {What determines the order of verbal dependents in Hindi?
  Effects of efficiency in comprehension and production} {What determines the
  order of verbal dependents in hindi? effects of efficiency in comprehension
  and production}.{\BBCQ}
\newblock
\BIn{} \APACrefbtitle {Proceedings of the Workshop on Cognitive Modeling and
  Computational Linguistics} {Proceedings of the workshop on cognitive modeling
  and computational linguistics}\ (\BPGS\ 1--10).
\newblock
\begin{APACrefURL} \url{https://aclanthology.org/2020.cmcl-1.1/}
  \end{APACrefURL}
\PrintBackRefs{\CurrentBib}

\bibitem [\protect \citeauthoryear {%
Simon%
}{%
Simon%
}{%
{\protect \APACyear {1982}}%
}]{%
simon1982models}
\APACinsertmetastar {%
simon1982models}%
\begin{APACrefauthors}%
Simon, H\BPBI A.%
\end{APACrefauthors}%
\unskip\
\newblock
\APACrefYearMonthDay{1982}{}{}.
\newblock
{\BBOQ}\APACrefatitle {Models of Bounded Rationality, vols. 1 and 2} {Models of
  bounded rationality, vols. 1 and 2}.{\BBCQ}
\newblock
\APACjournalVolNumPages{Economic Analysis and Public Policy, MIT Press,
  Cambridge, Mass}{}{}{}.
\PrintBackRefs{\CurrentBib}

\bibitem [\protect \citeauthoryear {%
Simon%
}{%
Simon%
}{%
{\protect \APACyear {1990}}%
}]{%
simon1990invariants}
\APACinsertmetastar {%
simon1990invariants}%
\begin{APACrefauthors}%
Simon, H\BPBI A.%
\end{APACrefauthors}%
\unskip\
\newblock
\APACrefYearMonthDay{1990}{}{}.
\newblock
{\BBOQ}\APACrefatitle {Invariants of human behavior} {Invariants of human
  behavior}.{\BBCQ}
\newblock
\APACjournalVolNumPages{Annual review of psychology}{41}{1}{1--20}.
\newblock
\begin{APACrefURL}
  \url{https://www.annualreviews.org/doi/10.1146/annurev.ps.41.020190.000245}
  \end{APACrefURL}
\PrintBackRefs{\CurrentBib}

\bibitem [\protect \citeauthoryear {%
Simon%
}{%
Simon%
}{%
{\protect \APACyear {1991}}%
}]{%
simon1991cognitive}
\APACinsertmetastar {%
simon1991cognitive}%
\begin{APACrefauthors}%
Simon, H\BPBI A.%
\end{APACrefauthors}%
\unskip\
\newblock
\APACrefYearMonthDay{1991}{}{}.
\newblock
{\BBOQ}\APACrefatitle {Cognitive architectures and rational analysis: Comment}
  {Cognitive architectures and rational analysis: Comment}.{\BBCQ}
\newblock
\BIn{} K.~VanLehn\ (\BED), \APACrefbtitle {Architectures for intelligence: The
  Twenty-second Carnegie Mellon Symposium on Cognition} {Architectures for
  intelligence: The twenty-second carnegie mellon symposium on cognition}\
  (\BPGS\ 25--39).
\newblock
\APACaddressPublisher{}{Lawrence Erlbaum Associates, Inc.}
\newblock
\begin{APACrefURL} \url{https://apps.dtic.mil/sti/citations/ADA219199}
  \end{APACrefURL}
\PrintBackRefs{\CurrentBib}

\bibitem [\protect \citeauthoryear {%
Temperley%
}{%
Temperley%
}{%
{\protect \APACyear {2007}}%
}]{%
Temperley2007}
\APACinsertmetastar {%
Temperley2007}%
\begin{APACrefauthors}%
Temperley, D.%
\end{APACrefauthors}%
\unskip\
\newblock
\APACrefYearMonthDay{2007}{}{}.
\newblock
{\BBOQ}\APACrefatitle {Minimization of dependency length in written {E}nglish}
  {Minimization of dependency length in written {E}nglish}.{\BBCQ}
\newblock
\APACjournalVolNumPages{Cognition}{105}{2}{300--333}.
\newblock
\begin{APACrefURL}
  \url{http://www.sciencedirect.com/science/article/B6T24-4M7CDMS-2/2/e095449f6439b30003822a5838e53786}
  \end{APACrefURL}
\newblock
\begin{APACrefDOI} \doi{DOI: 10.1016/j.cognition.2006.09.011} \end{APACrefDOI}
\PrintBackRefs{\CurrentBib}

\bibitem [\protect \citeauthoryear {%
Temperley%
\ \BBA {} Gildea%
}{%
Temperley%
\ \BBA {} Gildea%
}{%
{\protect \APACyear {2018}}%
}]{%
temperley-gildea-ar18}
\APACinsertmetastar {%
temperley-gildea-ar18}%
\begin{APACrefauthors}%
Temperley, D.%
\BCBT {}\ \BBA {} Gildea, D.%
\end{APACrefauthors}%
\unskip\
\newblock
\APACrefYearMonthDay{2018}{}{}.
\newblock
{\BBOQ}\APACrefatitle {Minimizing Syntactic Dependency Lengths:
  Typological/Cognitive Universal?} {Minimizing syntactic dependency lengths:
  Typological/cognitive universal?}{\BBCQ}
\newblock
\APACjournalVolNumPages{Annual Reviews of Linguistics}{}{}{}.
\newblock
\begin{APACrefURL}
  \url{https://www.cs.rochester.edu/u/gildea/pubs/temperley-gildea-ar18.pdf}
  \end{APACrefURL}
\PrintBackRefs{\CurrentBib}

\bibitem [\protect \citeauthoryear {%
Yamashita%
\ \BBA {} Chang%
}{%
Yamashita%
\ \BBA {} Chang%
}{%
{\protect \APACyear {2001}}%
}]{%
yamashitaChang2001}
\APACinsertmetastar {%
yamashitaChang2001}%
\begin{APACrefauthors}%
Yamashita, H.%
\BCBT {}\ \BBA {} Chang, F.%
\end{APACrefauthors}%
\unskip\
\newblock
\APACrefYearMonthDay{2001}{}{}.
\newblock
{\BBOQ}\APACrefatitle {{``Long before short'' preference in the production of a
  head-final language}} {{``Long before short'' preference in the production of
  a head-final language}}.{\BBCQ}
\newblock
\APACjournalVolNumPages{Cognition}{81}{}{}.
\newblock
\begin{APACrefURL}
  \url{https://www.sciencedirect.com/science/article/pii/S0010027701001214}
  \end{APACrefURL}
\PrintBackRefs{\CurrentBib}

\end{thebibliography}

\end{document}